# Question-Answering (QA) Model for a Personalized Learning Assistant for Arabic Language

Mohammad Sammoudi[1], Ahmad Habaybeh[1], Huthaifa I. Ashqar[2], and Mohammed Elhenawy[3]

[1] Department of Natural, Engineering and Technology Sciences, Arab American University, Ramallah P.O. Box 240, Palestine
[2] Civil Engineering Department, Arab American University, Jenin P.O. Box 240, Palestine and Columbia University, NY, USA
`Huthaifa.ashqar@aaup.edu`
[3] CARRS-Q, Queensland University of Technology, Brisbane, Australia

**Abstract.** This paper describes the creation, optimization, and assessment of a question-answering (QA) model for a personalized learning assistant that uses BERT transformers customized for the Arabic language. The model was particularly fine-tuned on science textbooks in Palestinian curriculum. Our approach uses BERT's brilliant capabilities to automatically produce correct answers to questions in the field of science education. The model's ability to understand and extract pertinent information is improved by fine-tuning it using 11th and 12th grade biology book in Palestinian curriculum. This increases the model's efficacy in producing enlightening responses. Exact match (EM) and F1 score metrics are used to assess the model's performance; the results show an EM score of 20% and an F1 score of 51%. These findings show that the model can comprehend and react to questions in the context of Palestinian science book. The results demonstrate the potential of BERT-based QA models to support learning and understanding Arabic students' questions.

## 1 Introduction

Wars, strikes and pandemics cause and create a gap between the students and their teachers. The students are having their book (contexts) and can self-study these books. However, when they are studying, they face questions and challenges that require a help.

The use of technology in the field of education has become essential to improve learning experiences in the constantly changing education methods. The developing of learning assistant systems, especially those suited for classroom use, is one potential direction. These systems use Question-Answer (QA) and Natural Language Processing (NLP) to build intelligent and interactive platforms that meet each student's unique demands [1].

A field of artificial intelligence called natural language processing gives computers the ability to comprehend, interpret, and produce language that is similar to that of humans. When NLP is included into educational systems, it makes it easier for students and educators to communicate smoothly, which promotes a more customized and interesting learning environment [2].

The Question-Answer model, which is the core of these systems, is made to efficiently understand and react to user inquiries. Because of the large amount of data that this model has been trained on, it can recognize linguistic nuances and offer precise, contextually appropriate answers. Learning Assistant Systems can serve as virtual tutors, helping students understand difficult ideas, work through issues, and reinforce what they learn in the classroom by putting QA models into action [3].

A system like this is important because it can adjust to the different learning styles and speeds of each learner. Learning Assistant Systems, fueled by NLP and QA models, provide a dynamic and customized learning experience. These tools may adjust to meet the particular requirements of every learner, offering tailored assistance and direction. In order to explain how NLP and QA models work together to create an intelligent and engaging teaching aid, this paper examines the design and implementation of a learning assistant system for students in schools.

## 2    Literature Review

Many studies have studied the idea of personalized learning assistant (PLA). These studies aim to build models that can help students in their study and examination [4]. Most of these studies used Natural Language Processing (NLP) techniques to extract information from students and support them with suitable answers, exercises, and materials [5].

The process of extract information from users to understand their query and give them an answer is the main challenge. A lot of models are used for this purpose. In [6] TF-IDF was used to match questions with similar questions that are found on search engines, this was a step in the right direction but a problem is faced when some question is not present on Internet, so we will not get an answer.

NLP tasks can be used for any language, but you need to be careful to use the version of model which support the desired language. For example, a study by Yilmaz and Toklu use NLP with Turkish language for analysis on question classification task. They use Word2Vec method with skip gram on a large corpus consists of user questions. They also use Convolutional Neural Networks (CNN) for classification task [7].

The study [8] explored the usage of chatbots in the pedagogical setting between 2017 and 2021. It evaluated the literature concerning the implementation of chatbots in education. The findings encompass the advantages and disadvantages of employing chatbots in e-learning, strategies to enhance and address the drawbacks, and the role of personalization in student advancement.

A paper [9] introduces the idea of interactive agents, such as Siri and Alexa, and their usage in everyday life, specifically in the context of higher education. It highlights the potential use of these agents as digital tutors. The paper concludes by analyzing the creation of conversational agents and their effect on learning outcomes.

A project explained in [10] suggest the use of AI and NLP techniques to generate personalized feedback automatically, including hints and explanations based on Wikipedia, without the need for expert intervention. This approach aims to improve student performance outcomes and is demonstrated through the implementation of the Korbit

43ITS, which serves approximately 20,000 students. The strategy employed entails utilizing ML and NLP techniques to create tailored feedback.

Chatbots as PLAs was used a lot in this field. For example, A study [11] explores the fusion of AI and NLP in the design of an ITS for school education, particularly focusing on computer science subjects. In this study a chatbot was developed and they conclude that chatbots have significant potential in the realm of school education, particularly if enhancements are made to the knowledge base and language models.

Another chatbot called LANA, proposed in [12], is a Conversational Intelligent Tutoring System (CITS) specifically designed for Arabic-speaking children with Autism Spectrum Disorder (ASD) who exhibit high-functioning autism or Asperger's Syndrome. LANA aims to provide a personalized learning experience in Arabic for individuals aged 10 to 16 by leveraging the structured nature of technology, which is often preferred by individuals with ASD.

An AI learning assistant called HAnS is the outcome of a collaboration between German universities and research institutes. It incorporates Natural Language Processing (NLP), speech recognition, and AI-based indexing to promote self-directed and personalized learning [13].

All the above studies use unidirectional techniques, which have many constraints. The inherent recurrent structure of these models makes it difficult to handle parallel processing. Additionally, they struggle with long sentences due to the vanishing gradient problem [14].

In 2017, Vaswani et al. proposed a new neural network model named Transformer [15]. This model introduced a modern architecture that inspired new developments like BERT and GPT, which rely heavily on the Transformer architecture. GPT is an autoregressive model and is considered a unidirectional language model [16]. In contrast, BERT provides a bidirectional representation, allowing for a better understanding of the text [17].

The implications of these advancements are significant. LANA and HAnS, while innovative, are limited by their use of unidirectional models, which face challenges in parallel processing and managing long sentences. The Transformer model, with its ability to handle parallel processes and mitigate the vanishing gradient problem, represents a major leap forward. This has led to the development of BERT and GPT models, which offer improved performance. BERT, with its bidirectional representation, enhances contextual understanding, making it particularly powerful for tasks requiring nuanced text comprehension. GPT, being autoregressive, excels in text generation tasks, but its unidirectional nature poses certain limitations in understanding context compared to BERT. These insights highlight the importance of selecting appropriate model architectures to address specific challenges in AI and NLP applications.

In our study, we focus on using a pre-trained model called BERT and fine-tuning it with a dataset containing Arabic texts from the Palestinian curriculum biology books for 11th and 12th grades. This approach differentiates our work from other models described in the literature review in several significant ways.

Firstly, while existing systems like LANA and HAnS have made strides in providing personalized learning experiences, they typically use unidirectional models or focus on different user demographics and educational content. For instance, LANA is designed for Arabic-speaking children with Autism Spectrum Disorder, leveraging the structured nature of technology to support individuals aged 10 to 16. However, LANA does not specifically utilize a pre-trained BERT model fine-tuned with educational content from a specific curriculum.

Secondly, the HAnS assistant integrates NLP, speech recognition, and AI-based indexing to facilitate self-directed learning but does not target Arabic texts from a specific educational curriculum. Additionally, HAnS is a product of collaboration between German institutions and is primarily designed with their context in mind, rather than focusing on Arabic educational content.

Our study's uniqueness lies in the use of BERT, a bidirectional transformer model that provides a deeper contextual understanding of the text. By fine-tuning BERT with Arabic texts from the Palestinian curriculum's biology books, we ensure that the model is tailored to the specific language nuances and educational needs of Arabic-speaking students in Palestine. This curriculum-specific fine-tuning enhances the model's relevance and accuracy in understanding and responding to educational queries related to biology for grades 11 and 12.

Moreover, our model is designed to predict suitable responses based on the student's level, input context, and query, providing a highly personalized learning experience. This contrasts with the generic or less specialized models discussed previously, which may not offer the same level of contextual understanding and specificity in their responses.

By focusing on fine-tuning BERT with a targeted Arabic dataset from the Palestinian curriculum, our work aims to fill a critical gap in educational AI tools, offering a solution that is not only linguistically and contextually relevant but also highly specialized for the intended educational content. This approach leverages the strengths of BERT's bidirectional representation to improve the precision and relevance of the responses, ultimately supporting more effective and personalized learning experiences for Arabic-speaking students.

## 3   Materials and Methods

For our research we follow the main NLP methodology which followed for almost every NLP Project. However, to make it suitable for our project, we modify it in some way to be suitable for our main goal. Our main project goal is to create a Personalized Learning Assistant (PLA) that will help school students. The first step was to choose a subject from the Palestinian curriculum and extract useful information from it. We choose the science subject for the school students for the secondary schools.

### 3.1   Data Collection and Preparation

To achieve the research goals two datasets were used, the first dataset was collected by a colleague team as a part of an NLP related project in a master's Natural Language Processing class held in Arab American University /Palestine in fall/2023. This Arabic dataset was manually collected from the 11th and 12th grade's biology books which are taught in Palestine, the dataset contains 111 observations and 7 attributes, the attributes are: id, unit_title, lesson_title, section_title, section_content , questions, avaliable_summary.

The second dataset which contains the student information is a publicly available dataset on www.kaggle.com (https://www.kaggle.com/datasets/aljarah/xAPI-Edu-Data). The origin of the dataset is The University of Jordan, Amman, Jordan. It was collected by Elaf Abu Amrieh, Thair Hamtini, and Ibrahim Aljarah from learning management system (LMS) called Kalboard 360. The dataset consists of 480 student records and 16 features, table 1 shows features descriptions and distinct values of each feature. The students in the dataset are classified into three categories: Low-Level 0 to 69, Middle-Level 70 to 89, High-Level 90 to 100.

Table 1. Students Profiles Dataset.

| Id | Feature name | Description | Data Type | Available values |
|---|---|---|---|---|
| ID | Feature Name | Description | Data Type | Available Values |
| 1 | Gender | Student's Gender | Nominal | Male, Female |
| 2 | Nationality | Student's Nationality | Nominal | Kuwait, Lebanon, Egypt, SaudiArabia, USA, Jordan, Venezuela, Iran, Tunis, Morocco, Syria, Palestine, Iraq, Lybia |
| 3 | Place of birth | Student's Place of Birth | Nominal | Kuwait, Lebanon, Egypt, SaudiArabia, USA, Jordan, Venezuela, Iran, Tunis, Morocco, Syria, Palestine, Iraq, Lybia |
| 4 | Educational Stages | Educational Level Student Belongs | Nominal | Lowerlevel, MiddleSchool , HighSchool |
| 5 | Grade Levels | Grade Student Belongs | Nominal | G-01, G-02, G-03, G-04, G-05, G-06, G-07, G-08, G-09, G-10, G-11, G-12 |
| 6 | Section ID | Classroom Student Belongs | Nominal | A, B, C |
| 7 | Topic | Course Topic | Nominal | English, Spanish, French, Arabic, IT, Math, Chemistry, Biology, Science, History, Quran, Geology |
| 8 | Semester | School Year Semester | Nominal | First, Second |
| 9 | Parent responsible for student | Parent responsible for student | Nominal | Mom, Father |

## 3.2  Problem Statement and Proposed Framework

During In recent years, Palestine has faced significant challenges, including wars and financial difficulties, leading to frequent teachers' strikes. These disruptions have created a gap between students and their educators, making it difficult for students to receive the necessary guidance and support. Despite having access to their science books from the Palestinian curriculum, students often struggle to find the answers to their questions without teacher assistance.

To address this issue, we propose a model designed to retrieve answers for students' questions based on the context provided in their science books. Our model is fine-tuned specifically for the Arabic language using texts from the Palestinian curriculum's science books. This fine-tuning ensures that the model understands the specific linguistic and educational context of Palestinian students.

Our model serves as a learning assistant, leveraging advanced NLP techniques to provide accurate and contextually relevant answers. By incorporating personalization, the model tailors its responses based on the student's educational level, ensuring that the answers are appropriate for their understanding and needs. This personalized approach helps bridge the gap left by the absence of teachers, offering students a reliable resource to continue their education independently.

Using a pre-trained BERT model, fine-tuned with domain-specific content, we enhance the model's ability to comprehend and respond to queries effectively. This method not only improves the accuracy of the answers but also ensures that they align closely with the curriculum, providing a seamless learning experience. The deployment of this model supports continuous learning despite external challenges, empowering students to progress in their education autonomously.

Our solution also alleviates the pressure on the educational system by providing an alternative means of instruction during times when teachers are unavailable. By ensuring students have access to accurate and relevant information, we contribute to maintaining educational standards and continuity. This initiative highlights the potential of AI-driven solutions in overcoming educational barriers and enhancing the learning experience in challenging environments.

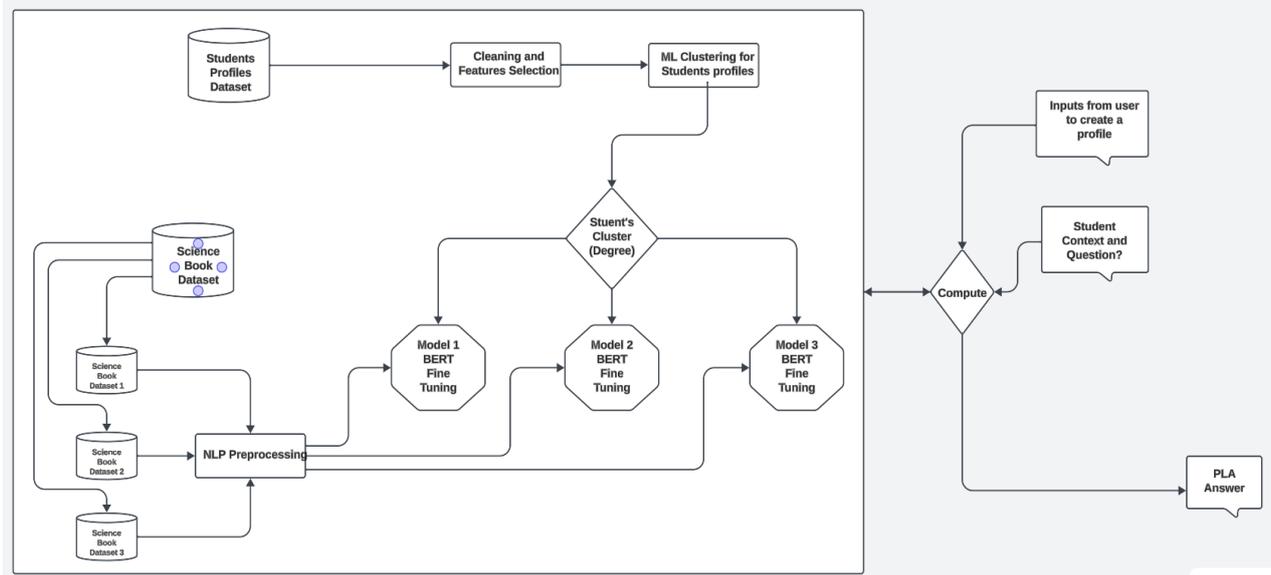

**Fig. 1.** Project Implementation Flowchart.

The above flowchart in Fig. 1 explains the methodology followed in our project. The project is consisting of two stages: the first stage is student's personalization and the second is the model building and fine tuning.

**Students Personalization.** The student's profiles dataset is used for this part of project. We get this dataset form Kaggle. Next, we clean the dataset by deleting any NA values and ensure the dataset is cleaned and all features' data types are suitable to be used by the machine learning model. Then, we did a classification task using K-Means clustering unsupervised ML model. We used the elbow method to find the optimal number of clusters, we found using elbow method that the students are distributed among three clusters, we will have three students' profiles. The three clusters that we have are classified based on their behavior (Excellent students, Good students and weak students).

**Question-Answer Bert Fine Tuning Model.** The second part of our project will be fine tuning BERT model to get a suitable answer for the student inquires. We will use a dataset collected from the Palestinian science book which is collected by our colleagues.

The main dataset is subdivided into three datasets. Each one will be different from the other by the question part which will be suitable for the cluster students from the first part of the model. We then fine-tuned the BERT model for Question-Answering task one for each cluster.

We used a pretrained BERT model for question answering (base-multilingual-cased). This model was created by Google and fine-tuned on XQuAD like data for 11 different languages including Arabic. Number of parameters in Bert model is 100 M [6].

**User Interaction with the model.** When the user interacts and use the model, he will be asked to input some information for the purpose to create his profile and for user classification later.

Then user will put context from the Palestinian science book. After that, he will ask a question to the model.

The model will use BERT fine-tuned model with the suitable dataset for the cluster that the student belongs to and gives an answer for the student.
We will use the BERT-base-multilingual-cased and fine-tuned it with our datasets.

## 4  Results and Discussion

The proposed framework illustrates a comprehensive process for tailoring educational content to students based on their profiles using NLP techniques. The process begins with a dataset of students' profiles, which undergoes cleaning and feature selection to prepare the data for clustering. This step ensures that the data is accurate and that only the most relevant features are used. The cleaned data is then processed using ML algorithms to cluster students into groups based on their profiles, categorized by degree. Concurrently, datasets from science books (Dataset 1, Dataset 2, and Dataset 3) are subjected to NLP preprocessing to extract relevant information and prepare the text data for model training. Each student cluster is mapped to a specific BERT model that has been fine-tuned on these preprocessed datasets. This fine-tuning allows the models to provide answers tailored to the context of each student cluster. New inputs from users (i.e., students) are collected to create or update their profiles, ensuring the system has current and accurate data. The system considers the student's context and specific questions, and based on this information, selects the appropriate BERT model to compute the most relevant answer. The final output is a PLA answer that is specifically tailored to the student's needs, leveraging advanced ML and NLP techniques to enhance the learning experience. This framework ensures that educational content is personalized and relevant, thereby optimizing the learning process for each individual student.

We used a QA system using multilingual BERT model. This model is trained on SQuAD v1 dataset. We fine-tuned this model using our dataset in Arabic language for the science book in Palestinian curriculum. Quantifying the success of question answering model is not straight and can be tricky. This is because when you ask a question to multiple people, each of them will answer differently from others. The answers may be the same in context but mostly they differ in synonyms or the order of words.

There are two primary evaluation metrics for question answering models. The first one is Exact Match (EM) and the other is F1-Score. The overall EM and F1 scores are computed by taking the average over the individual example scores.

EM is simple and very clear. It compares the predicted answer obtained from the model with the true answer found in the dataset. Its value will be either 1 if an exact match is found between the two answers, and 0 if no match is found. The model we fine-tuned uses the Arabic language, and as we know, Arabic words are very diverse, leading to a lower expectation for the EM metric. We used the Arabic SQuAD dataset to evaluate our model. Specifically, we took 693 question-answer pairs from the dataset, then used only the questions to predict the answers. The average EM for our model predictions is 20%, which means that our model is able to provide the exact answer 20% of the time.

The implications of a 20% EM score highlights the inherent challenges in processing and understanding Arabic text, given the language's morphological richness and syntactic complexity. This relatively low EM score suggests that while the model has some capability to generate accurate answers, there is significant room for improvement. This could be due to various factors, such as the quality and size of the training data, the model architecture, or the preprocessing steps used. Furthermore, a 20% EM indicates that the model might be generating plausible but not exact answers for a large portion of the questions, which can still be useful in many practical applications where an approximate answer suffices. However, for applications requiring high precision and exact matches, additional efforts in model refinement, such as incorporating more diverse training data, enhancing the model's understanding of context, or using more advanced NLP techniques, are necessary. This evaluation underscores the importance of continuous iteration and enhancement in developing robust NLP models for complex languages like Arabic.

F1 score is also a common metric for all classification problems. It depends on recall and precision. F1 score can be found using the following formula:

$$F1 = 2 \times \frac{\text{Precision} \times \text{Recall}}{\text{Precision} + \text{Recall}}$$

In our case, the F1 score is computed by finding the probabilities of individual words in the prediction against words in the true answer. The number of shared words between the predicted answer and the true answer is used to calculate precision, recall, and the F1-score. Precision is the ratio of the number of shared words to the total number of words in the prediction, while recall is the ratio of the number of shared words to the total number of words in the true answer. The average F1 score for our model predictions, based on 693 question-answer pairs, is 51%, which is considered good for a model using the Arabic language.

The implications of a 51% F1 score are significant. This metric, which balances precision and recall, provides a more nuanced evaluation of the model's performance than the EM score alone. An F1 score of 51% indicates that the model is fairly effective in capturing the essence of the true answers, even if it does not always produce exact matches. This is particularly impressive given the complexities and variations in the Arabic language, including its rich morphology, diverse dialects, and complex syntactic structures.

A 51% F1 score suggests that the model can generate answers that are partially correct and contextually relevant, even if they do not match the true answers word-for-word. This level of performance is useful in real-world applications where an exact match is not always necessary, and approximate correctness is acceptable. For example, in information retrieval and question-answering systems, providing a relevant and contextually accurate answer can be more valuable than an exact match.

However, this also indicates that there is room for improvement. To further enhance the model's performance, additional training data, especially more diverse and representative datasets, could be used. Advanced preprocessing techniques and model architectures might also be explored to better capture the nuances of the Arabic language. Moreover, continuous model evaluation and refinement are essential to gradually increase both the EM and F1 scores, making the model more robust and accurate.

## 5      Conclusion

The building of personalized learning assistant and use of a question-answering model for school science textbooks and Arabic language that makes use of BERT transformers offers encouraging developments in natural language processing in educational environments. We have shown through this study that it is feasible and efficient to use pretrained language models such as BERT to automatically produce correct answers to questions in Arabic, especially in the field of science education in Palestinian schools. Moreover, incorporating this kind of technology has a great deal of promise to help Arabic-speaking students to learn and understand. This concept can be a useful tool for teachers and students, providing students with immediate access to precise explanations and clarifications that are customized to the content of school science books and personalized to their performance level.

The effective creation and assessment of a learning assistant BERT-based question answering model for Arabic language and school science textbooks highlights the new advantages of NLP in learning environments. We may improve Arabic-speaking students' ability to access to high-quality educational resources and allow them to benefit from AI and NLP techniques. This will ultimately contribute to the growth of education in this digital age.